\documentclass{article}
\usepackage{amsmath,graphicx,mlspconf,amssymb}
\usepackage{physics}
\usepackage{multirow}
\usepackage{accents}
\usepackage[ruled,vlined]{algorithm2e}
\usepackage{algorithmic}
\usepackage{xcolor}
\DeclareMathOperator*{\argmax}{arg\,max}

\newcommand\munderbar[1]{%
  \underaccent{\bar}{#1}}
\copyrightnotice{978-1-7281-6662-9/20/\$31.00 {\copyright}2020 IEEE}

\toappear{2020 IEEE International Workshop on Machine Learning for Signal Processing, Sept.\ 21--24, 2020, Espoo, Finland}

\title{Robust classification using hidden Markov models and mixtures of
normalizing flows}
\name{Anubhab Ghosh, Antoine Honor\'e, Dong Liu, Gustav Eje Henter, Saikat Chatterjee}
\address{School of Electrical Engineering and Computer Science, KTH Royal Institute of Technology, Sweden}

\RequirePackage[normalem]{ulem} 
\RequirePackage{color}\definecolor{RED}{rgb}{1,0,0}\definecolor{BLUE}{rgb}{0,0,1} 

\begin{document}

\maketitle

\begin{abstract}

We test the robustness of a maximum-likelihood (ML) based classifier where sequential data as observation is corrupted by noise. The hypothesis is that a generative model, that combines the state transitions of a hidden Markov model (HMM) and the neural network based probability distributions for the hidden states of the HMM, can provide a robust classification performance. The combined model is called normalizing-flow mixture model based HMM (NMM-HMM). 
It can be trained using a combination of expectation-maximization (EM) and backpropagation. We verify the improved robustness of NMM-HMM classifiers in an application to speech recognition.
\end{abstract}
\section{Introduction}
\label{sec:intro}

Neural networks have received much attention in the last decade for their success in regression and pattern classification. Discriminative systems based on neural networks have often been found to outperform many classical classification methods. The systems are data-driven and model-free. The power of neural networks is in non-linear signal transformations through multiple layers. An educated guess is that the non-linear transformations produce appropriate features for  classification. While explored as an active research area, neural networks and deep architectures have been criticised for low explainability.
We will provide a clear example where neural networks have lack explainability.

Discriminative classification setups optimize a cost function that directly uses a discrimination measure between classes to achieve a high classification performance. It is used to answer pre-fixed queries. The number of classes and class labels are fixed in the optimization and prediction as well.
The optimization is known as `discriminative training'. 

Neural network based discriminative classification methods are known to suffer from robustness issues. It is found that a small, often imperceptible amount of noise, in the input signal / features can lead to wrong classifications. Small perturbations known as `adversarial perturbations', can be engineered to confuse a neural network based discriminative classifier \cite{su2019one}, \cite{moosavi2016deepfool}. The reason for this lack of robustness is not fully understood. This is a clear example of where we lack explanations. 

We perceive that data-driven, model-free, discriminative classification systems are hard to analyze. Lack of analysis with mathematical tractability hinders finding the reasons for failures or methods to mitigate the issue.   

Model-based systems can be amenable for scrutiny and tractable mathematical analysis, which is a basic precursor of explanations. We propose to combine the analytical tractability of model-based systems and the non-linear transformation advantage of neural networks. The combination will provide a better classification performance than a set of pure model-based systems. The combined models will provide robustness. Robustness can be experimentally verified using real-world data and various types of noises. In this article our main objective is to provide such an experimental verification.

In pursuit of our objective, we use speech recognition as a real application scenario. Hidden Markov models (HMMs) are time-tested in speech recognition \cite{gales2008application}. HMMs can model sequential data and have been used in a variety of applications including handwriting recognition, activity recognition, genetic signal processing, transport forecasting, etc. For speech recognition, probability distributions of states in an HMM are classically modelled using Gaussian mixture models (GMMs). The GMM-HMM combination is purely model-driven. GMM-HMMs are generative models, meaning they can explain how sequential data is generated. It can be trained using time-tested probabilistic methods, such as expectation-maximization (EM), variational Bayes (VB) and Markov-Chain-Monte-Carlo (MCMC). After training, GMM-HMM models can be used directly for ML based classification.

ML based classification using generative models addresses Bayes minimum risk criterion. It can easily accommodate a growing number of classes. Adversarial perturbation is not relevant in the case where ML-based classification is performed using generative models. However, robustness for various types of noises remains an important issue.

Recent advances have improved the ability of the neural network based models to describe probability distributions. EM was used for training mixture models based on 'normalizing-flow' in \cite{liu2020explicit}. Normalizing-flow is a generative model \cite{Dinh2015}, \cite{kobyzev2019normalizing}. Normalizing-flow mixture models (NMMs) are also generative models and may handle multiple modes and manifolds in a data distribution better than GMMs. In \cite{liu2019powering}, NMMs were used for modelling the state distributions in an HMM. In this article, we call this model as NMM-HMM in contrast to GMM-HMM. NMM-HMM parameters can be learned using EM and backpropagation. Our main technical contribution in this article is to show than the NMM-HMM is more robust than GMM-HMM for speech recognition at various noisy conditions in ML-based classification. We verify this in an extensive experimental study of robust phone recognition on the TIMIT database using the Kaldi and PyTorch toolkits.
\subsection{Relevant literature survey for speech recognition}
Phonemes can be thought of as one of the fundamental units of speech sounds, that indicate the pronunciation of a word. A spoken utterance, which is essentially a sequence of words, is composed of phonemes. The task of phone recognition involves creating a system that can be given input as a human spoken utterance, and the system outputs a sequence of phones that indicate the pronunciation of the actual spoken text. Speech phone recognition is popular in the research community because it is a fundamental task in developing a speech recognition system \cite{lopes2011phone}. It is also inherently free from the limitations of vocabulary. The phone recognition accuracy continues to improve \cite{werweare}.


Neural networks are recently much used for speech recognition, mainly in discriminative training setups. A list of neural network based methods and their performances for phone recognition can be found in the Github site \cite{werweare}. An earlier attempt is to use  restricted Boltzmann machines (RBMs) to form a deep belief network (DBN) that served as the acoustic model for HMM \cite{mohamed2009deep}. 
The DBN used a softmax output layer and was trained discriminatively using backpropagation, achieving a phone recognition accuracy of around 77\% on the TIMIT dataset.

Dynamical system models have also been used for sequence-to-sequence classification. The dynamical neural networks are recurrent neural networks (RNNs), long-short-term-memory networks (LSTMs), and their gated recurrent unit based simplifications. Attention mechanisms can further improve performance. Example works are \cite{Graves2013}, \cite{Chorowski2015}. Almost all these example works use discriminative training. All these methods used a softmax function to represent the phone probability estimation. The best phone recognition accuracy result so far known using discriminative training is $85.1\%$ \cite{Ravanelli2018}. 
Finally, the authors in \cite{Ravanelli2019} proposed a hybrid HMM based on PyTorch-Kaldi toolkit, whose phone recognition accuracy is mentioned as $86.2\%$.

\section{NMM-HMM}
\label{sec:format}
Our interest in this article is ML-based classification. Let the number of classes be $C$. We denote the generative model for $c$'th class as $\mathbf{H}_c$. We write a data sequence $\mathbf{\munderbar{x}} = \left[\mathbf{x}_1, \mathbf{x}_{2}, \ldots \mathbf{x}_{T}\right]^{T}$ (where the superscript $T$ denotes transpose), where each $\mathbf{x}_{t} \in \mathbb{R}^{D}$ is the feature vector extracting from the signal at time $t$ and $T$ denotes the sequence length. Then the ML-classification problem is
\begin{equation}
    c^{\star} = \argmax_c p(\mathbf{\munderbar{x}} | \mathbf{H}_c)
\end{equation}
where $p()$ denotes likelihood. Our interest is to use NMM-HMM as $\mathbf{H}_c$ instead of GMM-HMM. 
The NMM-HMM system has been described in detail in \cite{liu2019powering}. They called their model 'Gen-HMM'. We rename Gen-HMM as NMM-HMM (Normalizing-flow Mixture Model based HMM) since we compare with GMM-HMM. We restate some essential theory in the remainder of the section about NMM-HMM for consistency and completeness.

\subsection{Central principle behind NMM-HMM}
The framework used for implementing the normalizing flow-based model is a Hidden Markov Model (denoted by 
$\mathbf{H}$). 
An HMM is basically composed of the following quantities: the set of hidden states of the HMM denoted by $\mathrm{S}$, the initial probability vector for the states of the HMM denoted by $\mathbf{q}$, the state-transition probability matrix denoted by $\mathbf{A}$, and the output probability distribution for each state $s$ denoted by $p\left(\mathbf{x}|s;\mathbf{\Psi}_{s}\right)$ and parameterised by a set of parameters $\mathbf{\Psi}_{s}$.
The probabilistic model of the NMM-HMM for each hidden state is a weighted mixture of $K$ density functions that is defined as:
\begin{equation}
    p\left(\mathbf{x}|s;\mathbf{\Psi}_{s}\right) = \sum_{k=1}^{K} \pi_{s,k} p\left(\mathbf{x}|s;\mathbf{\phi}_{s,k}\right)
\label{eqn1}
\end{equation}
In \eqref{eqn1}, the weights $\pi_{s,k}$ denote the probability of drawing a given component $k$ from a categorical distribution with $\pi_{s,k} = p\left(k | s;\mathbf{H}\right)$ and they satisfy $\sum_{k=1}^{K} \pi_{s,k} = 1$ for each s. The feature vector $\mathbf{x}$ is considered to be generated from a flow-based \textit{generator} function $\mathbf{g}_{s,k} : \mathbb{R}^{D} \to \mathbb{R}^{D}$, such that $\mathbf{x} = \mathbf{g}_{s,k}\left(\mathbf{z} \right)$, where $\mathbf{z}$ is a D-dimensional latent variable that is drawn from a known prior distribution. As we are considering normalizing flow models, $p\left(\mathbf{z}\right)$ is assumed to be a standard multivariate normal distribution. Assuming the function $\mathbf{g}_{s,k}$ is invertible, and the corresponding \textit{normalizing} function is $\mathbf{f}_{s,k}$ (or equivalently $\mathbf{g}^{-1}_{s,k}$), s.t. $\mathbf{z} = \mathbf{f}_{s,k}\left(\mathbf{x} \right)$, we have:
\begin{equation}
p\left(\mathbf{x}|s;\mathbf{\Phi}_{s,k}\right) = p_{s,k} \left(\mathbf{f}_{s,k}\left(\mathbf{x} \right)\right) \abs{ \det \left( \frac{\partial \mathbf{f}_{s,k}\left(\mathbf{x} \right) }{\partial \mathbf{x}} \right)}
\label{eqn2}    
\end{equation}
This equation shows that the density computation is exact and tractable. In practice, a logarithm is applied on both sides of \eqref{eqn2} to transform it into a sum, and the log-likelihood of the signal is calculated using the log-probability of the prior and the log-determinant of the Jacobian.
\subsection{Learning problem formulation in NMM-HMM}
The NMM-HMM model is intended to be used for modeling a sequential signal such as $\mathbf{\munderbar{x}}$, having an empirical distribution of the dataset as $\hat{d}\left( \mathbf{\munderbar{x}}\right)$ and the maximum likelihood maximization problem that is required to solve is:
\begin{equation}
    \argmax_{\mathbf{H}} \frac{1}{R} \sum_{r = 1}^{R} \log \left( p \left(\mathbf{\munderbar{x}}^{r};\mathbf{H} \right) \right)
\label{eqn9}
\end{equation}
where $r$ denotes the index of the sequential signal in the training data, $R$ denotes the total number of sequential input signals under consideration and $H$ denotes the HMM model under consideration from the possible hypothesis set of models. 
This problem can be solved using the well known Expectation-Maximization framework. The Expectation  step ("E-step") involves the calculation of the posterior probability distribution of hidden sequences $\mathbf{\munderbar{s}}$ and $\mathbf{\munderbar{k}}$ to obtain an expected value of the likelihood of the data sequence $\mathbf{\munderbar{x}}$ under the current model parameters $\mathbf{{H}^{old}}$.". This is shown as follows:
\begin{equation}
L\left(\mathbf{H} ; \mathbf{H}^{old}\right) = \mathbf{E}_{p\left( \mathbf{\munderbar{s}}, \mathbf{\munderbar{k}} | \mathbf{\munderbar{x}}; \mathbf{H}^{old}\right)} \log \left( p\left( \mathbf{\munderbar{x}}, \mathbf{\munderbar{s}},  \mathbf{\munderbar{k}}; \mathbf{H}\right) \right)
\label{eqn10}    
\end{equation}
The second step consists of the Maximisation step ("M-step") consists of finding the model $\mathbf{H}$ that maximizes the expected log-likelihood computed in \eqref{eqn10}. This can be decomposed into three separate maximization problems as in \eqref{eqn11}:
\begin{multline}
\max_{\mathbf{H}} L\left(\mathbf{H} ; \mathbf{H}^{old} \right) \\
= \max_{\mathbf{q}} L\left(\mathbf{q}; \mathbf{H}^{old} \right) + \max_{\mathbf{A}} L\left(\mathbf{A}; \mathbf{H}^{old} \right) + 
\max_{\mathbf{\Psi}} L\left(\mathbf{\Psi}; \mathbf{H}^{old} \right)
\label{eqn11}
\end{multline}
where,
\begin{equation}
L\left(\mathbf{q}; \mathbf{H}^{old}\right) = \mathbf{E}_{p\left( \mathbf{\munderbar{s}} | \mathbf{\munderbar{x}}; \mathbf{H}^{old}\right)} \log \left( p\left( s_{1} ; \mathbf{H}\right) \right)
\label{eqn12}    
\end{equation}
\begin{equation}
L\left(\mathbf{A}; \mathbf{H}^{old}\right) = \mathbf{E}_{p\left( \mathbf{\munderbar{s}} | \mathbf{\munderbar{x}}; \mathbf{H}^{old}\right)} \sum_{t=2}^{T} \log \left( p\left( s_{t} | s_{t-1} ; \mathbf{H}\right) \right)
\label{eqn13}    
\end{equation}
\begin{equation}
L\left(\mathbf{\Psi}; \mathbf{H}^{old}\right) = \mathbf{E}_{p\left( \mathbf{\munderbar{s}}, \mathbf{\munderbar{k}} | \mathbf{\munderbar{x}}; \mathbf{H}^{old}\right)} \log \left( p\left( \mathbf{\munderbar{x}}, \mathbf{\munderbar{k}} | \mathbf{\munderbar{s}} ; \mathbf{H}\right) \right)
\label{eqn14}    
\end{equation}
The maximisation problems in \eqref{eqn12}, \eqref{eqn13} can be solved using standard EM forward-backward algorithm to compute the posterior distribution effectively, explained in \cite{bishop2006pattern}. For solving the maximisation problem of the last equation \ref{eqn14}, we need to maximise the likelihood with respect to the mixture of weights and the set of parameters of the flow model. This would require the computation of the log-determinant for the flow model that is explained in the next section.

\subsection{Implementation of individual generator models}
Each generator function is realised as a flow model. It maps a given observation from the feature space to a latent space of the same dimension. Every mapping from layer to layer should be bijective (i.e.\ one-to-one and invertible) and log-determinant in the inverse direction should be easy to compute. The signal flow for a flow model having $L$ layers can be illustrated as follows:
\begin{equation} \label{eqn3}
    \mathbf{z}=\mathbf{h_0} \underset{\mathbf{f}_{s,k}^{[1]}}{\stackrel{\mathbf{g}_{s,k}^{[1]}}{\rightleftharpoons}} \mathbf{h_1} \underset{\mathbf{f}_{s,k}^{[2]}}{\stackrel{\mathbf{g}_{s,k}^{[2]}}{\rightleftharpoons}} \mathbf{h_2} 
    \underset{\mathbf{f}_{s,k}^{[3]}}{\stackrel{\mathbf{g}_{s,k}^{[3]}}{\rightleftharpoons}} \mathbf{h_3} \hdots 
    \underset{\mathbf{f}_{s,k}^{[L]}}{\stackrel{\mathbf{g}_{s,k}^{[L]}}{\rightleftharpoons}} \mathbf{h_L} = \mathbf{x}
\end{equation}
where $\mathbf{f}_{s,k}^{[l]}$ denotes the $l^{th}$ layer network of $\mathbf{f}_{s,k}$, and each such $\mathbf{f}_{s,k}^{[l]}$ is invertible. Flow models have been first proposed in \cite{Dinh2015}. Different flow model architectures may have different kinds of \textit{coupling} between two successive layers in the network or some other bijective mapping.
To illustrate a small section of the mapping from the data space to the latent space, let us consider the input feature at the $l^{th}$ layer denoted by $\mathbf{h}_{l}$. This feature is mapped to $\mathbf{h}_{l-1}$ using the function $\mathbf{f}_{s,k}$. At every layer the $D$-dimensional input feature is split into two parts. Let us assume the features are $\left[\mathbf{h}_{l,1:d}, \mathbf{h}_{l,d+1:D}\right]^{T}$ (where $d$ denotes the number of components in the first sub part). The relation is as follows:
\begin{equation}
\begin{aligned}
\mathbf{h}_{l,1:d} &= \mathbf{h}_{l-1, 1:d} \\
\mathbf{h}_{l,d+1:D} &= \mathbf{h}_{l-1, d+1:D} \odot \exp\left(\mathbf{s} \left(\mathbf{h}_{l-1,1:d} \right) \right) + \mathbf{t} \left(\mathbf{h}_{l-1,1:d}\right)
\end{aligned}
\label{eqn4}
\end{equation}
The inverse function $(\mathbf{f} : X \to Z)$ (from data space to latent space, which is used in Jacobian computation during training) is defined as:
\begin{equation}
\begin{aligned}
\mathbf{h}_{l-1,1:d} &= \mathbf{h}_{l,1:d} \\
\mathbf{h}_{l-1,d+1:D} &= \left( \mathbf{h}_{l, d+1:D} - \mathbf{t} \left(\mathbf{h}_{l,1:d} \right) \right) \odot \exp\left(-\mathbf{s} \left(\mathbf{h}_{l,1:d}\right) \right) \\
\end{aligned}
\label{eqn5}
\end{equation}
(where $\odot$ denotes element-wise multiplications, $\mathbf{s}: \mathbb{R}^{d} \to \mathbb{R}^{D-d}$, $\mathbf{t}: \mathbb{R}^{d} \to \mathbb{R}^{D-d}$, with $\mathbf{s, t}$ being shallow feed-forward Neural Nets that differ only in the activation function for the last layer, which is a hyperbolic tangent (\textbf{\texttt{tanh}}) activation function for the scale function $\left(\mathbf{s}\right)$ (modeling logarithm of standard deviation) and an identity activation for the translation function $\left(\mathbf{t}\right) $\cite{Dinh2019}. For the flow model, the inverse mapping shown in \eqref{eqn5} is computed for every layer and the determinant of the Jacobian matrix is computed as the product of the determinants of the Jacobian matrices at every layer:
\begin{equation}
    \det\left(\nabla \mathbf{f}_{s,k} \right) = \prod_{l=1}^{L} \det\left(\nabla \mathbf{f}_{s,k}^{[l]} \right)
\label{eqn6}
\end{equation}
where each $\det\left(\nabla \mathbf{f}_{s,k}^{[l]} \right)$ is computed as:
\begin{equation}
\begin{aligned}
\det\left(\nabla \mathbf{f}_{s,k}^{[l]} \right) &= \det \left(\begin{bmatrix}
                                                \mathbf{I}_{1:d} & \mathbf{0} \\
                                                \frac{\partial \mathbf{h}_{l-1,d+1:D}}{\partial \mathbf{h}_{l-1,1:d}} & diag\left(-\mathbf{s}\left(\mathbf{h}_{l,1:d}\right)\right)
                                                \end{bmatrix}
                                                \right) \\
                                    &= \det \left(diag\left(-\mathbf{s}\left(\mathbf{h}_{l,1:d}\right)\right) \right) \\
\end{aligned}
\label{eqn7}
\end{equation}
In \eqref{eqn7}, $\mathbf{I}_{1:d}$ denotes an identity matrix and $diag(\dotsc)$ denotes a diagonal matrix with the elements of the vector in the main diagonal. 
 It should be kept in mind and an alternate ordering between the two parts of the signal $\left[\mathbf{h}_{l,1:d}, \mathbf{h}_{l,d+1:D}\right]^{T}$, described in the affine coupling layer in \eqref{eqn4}, is required to avoid the problem of partial identity mapping \cite{Dinh2019} in the model. 
 For solution of \eqref{eqn14}, the problem can be broken down into two parts: learning the set of mixture of weights: $\mathbf{\Pi} = \lbrace \mathbf{\pi_{s}} | s \in S \rbrace $, and the learning the set of flow model parameters: $\mathbf{\Phi} = \lbrace \mathbf{\phi_{s}} | s \in S \rbrace $. This is shown as:
\begin{equation}
\max_{\mathbf{\Psi}} L\left(\mathbf{\Psi} ; \mathbf{H}^{old} \right) \\
= \max_{\mathbf{\Pi}} L\left(\mathbf{\Pi}; \mathbf{H}^{old} \right) + \max_{\mathbf{\Phi}} L\left(\mathbf{\Phi}; \mathbf{H}^{old} \right)
\label{eqn15}    
\end{equation}
The problem of learning the mixture of weights can be solved using a simple lagrangian formulation while problem of learning the flow model parameters can be solved by using the results of the change of variable \eqref{eqn2} and the log-determinant derived in \eqref{eqn7} as:
\begin{multline}
L\left(\mathbf{\Phi}; \mathbf{H}^{old} \right) = \mathbf{E}_{p\left( \mathbf{\munderbar{s}}, \mathbf{\munderbar{k}} | \mathbf{\munderbar{x}}; \mathbf{H}^{old}\right)} \log \left( p\left( \mathbf{\munderbar{x}} | \mathbf{\munderbar{s}} , \mathbf{\munderbar{k}} ; \mathbf{H}\right) \right) \\
= \mathbf{E}_{p\left(\mathbf{\munderbar{s}}, \mathbf{\munderbar{k}} | \mathbf{\munderbar{x}}; \mathbf{H}^{old} \right)} [ \log \left( p_{\mathbf{\munderbar{s}},  \mathbf{\munderbar{k}}} \left( \mathbf{f}_{s,k}\left(\mathbf{\munderbar{x}} \right) \right) \right) \\
+ \log \left(\abs{ \det \left(\nabla \mathbf{f}_{s,k} \right)} \right) ]
\label{eqn16}
\end{multline}
The first term on the right hand side of \eqref{eqn16} is basically an expectation computed over the log-probability of latent data derived using the normalizing function $\mathbf{f}_{s,k}: X \to Z$, and the second term is the result of the log-determinant of the Jacobian that is computed using  \eqref{eqn6} and \eqref{eqn7}. The essential steps of the learning procedure are described in psuedocode in Algorithm \ref{algo1}.
\begin{algorithm}
\SetAlgoLined
 \KwIn{Dataset $\mathbf{\munderbar{x}}$, initial model parameters and the empirical distribution of the dataset $\hat{d}\left( \mathbf{\munderbar{x}}\right)$}
\KwResult{Optimized model parameters: $\mathbf{q}, \mathbf{A}, \mathbf{\Pi}, \mathbf{\Phi}$}
Set learning rate $(\eta)$, no. of mini-batches, max. epochs for flows $(N_{max})$ \\
Initialization of $\mathbf{H^{old}}$, $\mathbf{H}$:
where, $\mathbf{H} = \lbrace \mathbf{q}, \mathbf{A}, S, p\left(\mathbf{x}|s;\mathbf{\Psi}_{s}\right) \rbrace$,
and \\
set $\mathbf{H^{old}} \gets \mathbf{H}$\\
 \While{$\mathbf{H}$ has not converged}{
  \For{$\text{num\_epochs} \leq N_{max}$}{
  Input a mini-batch from the dataset as $\lbrace \mathbf{\munderbar{x}}^{r} \rbrace ^{R_{b}}_{r = 1}$, with batch-size $R_{b}$ \\
  Compute the posterior $p\left( \mathbf{\munderbar{s}}^{r}, \mathbf{\munderbar{k}}^{r} | \mathbf{\munderbar{x}}^{r}; \mathbf{H}^{old}\right)$
  and the loss function $L\left(\mathbf{\Phi}; \mathbf{H}^{old} \right)$ shown in \eqref{eqn16} \\
  Compute $\partial \mathbf{\Phi}$ by optimising $L\left(\mathbf{\Phi}; \mathbf{H}^{old} \right)$\\
  Update $\mathbf{\Phi} \gets \mathbf{\Phi} + \eta \cdot \partial \mathbf{\Phi}$\\
  }
  $\mathbf{q} \gets \argmax_{\mathbf{q}}L\left(\mathbf{q}; \mathbf{H}^{old} \right)$ \\
  $\mathbf{A} \gets \argmax_{\mathbf{A}}L\left(\mathbf{A}; \mathbf{H}^{old} \right)$ \\
  $\mathbf{\Pi} \gets \argmax_{\mathbf{\Pi}}L\left(\mathbf{\Pi}; \mathbf{H}^{old} \right)$ \\
  Update $\mathbf{H^{old}} \gets \mathbf{H}$
 }
 \caption{Learning Algorithm of NMM-HMM}
 \label{algo1}
\end{algorithm}

\section{Experiments and Results}
\label{sec:experiments}
This section describes the experimental setup for comparing the performance of the NMM-HMM model and the conventional GMM-HMM model, on the task of phone recognition. 
\subsection{Dataset and feature extraction}
The experiments were carried out using the TIMIT dataset \cite{lopes2011phone}, which has utterances labelled at the phoneme level. The speech signal in the dataset has been sampled at 16 kHz, and the dataset consists of 6300 phoneme-level speech utterances that have been split into two sets - A training set consisting of 4620 utterances and a testing set consisting of 1680 utterances. In the original TIMIT dataset there are 61 phones available for classification. There also exists a smaller, 'folded' set of 39 phones mapped from the larger set of phones \cite{lopes2011phone}. The experiments were performed for classification on the folded set of 39 phones. Four varieties of additive noises from from the NOISEX-92 database (also sampled at the same frequency of 16 kHz) were considered and snippets were taken from random offsets in the noise database and were added to the clean test data at different Signal to Noise ratio levels (with respect to the clean test signal) to generate noisy test sets. The features used were Mel frequency Cepstral Coefficients (MFCCs) that were extracted on a per frame basis (25 ms window frames, with a 10ms time shift between successive windows). This resulted in a vector of 13 MFCCs, along with column-wise concatenation of velocity ($\Delta$ coefficients) and acceleration coefficients ($\Delta\Delta$ coefficients) into a 39-dimensional feature vector. These feature vectors were subsequently processed, split into training and test sets and arranged on a per-class basis i.e. for each of the 39 phonemes present. This helped to set up the conditions for generative training of the models. 
\subsection{Model training}
The NMM-HMM models for each of the 39 classes of phones were trained using generative training. Training was carried out using Adam \cite{kingma2014adam} optimization, with data processed on a batch-wise basis. For each HMM model, the number of states used was clipped between 3 and 5, with the exact number of states used depending upon the mean sequence length of the incoming signal. For the GMM-HMM model, diagonal covariance matrices were considered for modeling each component Gaussian in the model, and the no. of mixture components were to be decided based on the basis of model performance. The state-transition matrix $\mathbf{A}$ for each NMM-HMM model was initialised as an upper triangular matrix. For each NMM-HMM model, a \textit{flow block} was used to refer to a pair of consecutive coupling layers that have been described in \eqref{eqn4},\eqref{eqn5}. It was necessary to ensure that the signal in the $l^{th}$ layer would be alternated by the ${(l+1)}^{th}$ layer, so that there is a mixing of the signals between consecutive coupling layers and no identity mapping occurs. Each NMM-HMM model consisted of \textit{four} such flow blocks in the implementation, which was found to work well. The evaluation was done using a full forward-backward algorithm, with the metric used as \textit{accuracy} $(100 - PER \%)$ computed on the test set, as a percentage of the number of correct predicted phonemes among the total set of phonemes. 
\subsection{Results}
\label{sec:results}

\subsubsection{Clean training and testing}
\label{sec:results1}
The performance of the NMM-HMM model was compared with the baseline GMM-HMM model for training and testing on clean data. For the purpose of comparison with the baseline model, the GMM-HMM model was trained and tested on the clean data, for varying number of mixture components, i.e. $K = \lbrace 3, 10, 15, 20 \rbrace$, and the model with the best performance was chosen for comparison with the NMM-HMM model. The results are shown in Table \ref{tab:GMM_diff_components}. It was found that accuracy increased with more number of mixture components. Based on the results, the model with $K=20$ components was chosen for comparison. 
\begin{table}[!t]
    \centering
    \caption{Recognition accuracy (in \%) for GMM-HMM at varying number of mixture components}
    
    \begin{tabular}{|*{5}{c|} }
            \hline
            \multirow{2}{*}{Model-Type}
            & \multicolumn{4}{c|}{No. of components (K)} \\
            \cline{2-5}
            & K=3 & K=10 & K=15 & K=20 \\
             \cline{1-5}
            GMM-HMM & 66.7 & 70.8 & 71.9 & 72.8 \\
            \hline
            \hline
        \end{tabular}
    \label{tab:GMM_diff_components}
\end{table}
Next we simulated NMM-HMM with varying number of mixture components. Results are shown in Table \ref{tab:NMM-HMM_diff_components}. It was found that $K=3$ components was best suited for the NMM-HMM model. There was some increase in accuracy by increasing number of components from $K=\lbrace 1, 3 \rbrace$, but with more number of components $(K > 3)$, the training time increased significantly without much change in test accuracy. Upon comparing Tables \ref{tab:GMM_diff_components} and \ref{tab:NMM-HMM_diff_components}, and find that the NMM-HMM model with $K=3$ mixture components outperformed the baseline GMM-HMM model with $K=20$ components. The performance improvement is $4.8\%$. 

We finally mention that the NMM-HMM based ML-classification provides a similar classification accuracy for TIMIT phone recognition in comparison with discriminative training based DBN that had $77\%$ accuracy \cite{mohamed2009deep}.  


\begin{table}[!t]
    \centering
    \caption{Recognition accuracy (in \%) for NMM-HMM at varying number of mixture components}
    
    \begin{tabular}{|*{4}{c|} }
            \hline
            \multirow{2}{*}{Model-Type}
            & \multicolumn{2}{c|}{No. of components (K)} \\
            \cline{2-3}
            & K=1 & K=3 \\
             \cline{1-3}
            NMM-HMM & 76.7 & 77.6 \\
            \hline
            \hline
        \end{tabular}
    
    \label{tab:NMM-HMM_diff_components}
\end{table}
\vspace{-1em}
\begin{table*}[!t]
    \centering
    \caption{Test accuracy (in \%) for clean and various noise conditions. We compare GMM-HMM and NMM-HMM for folded 39-phone classification. We use the notations GMM and NMM to represent GMM-HMM and NMM-HMM, respectively. The performance drop is shown in parenthesis with respect to the clean train and clean test scenario as in Tables \ref{tab:GMM_diff_components} and \ref{tab:NMM-HMM_diff_components}.}
    \setlength{\tabcolsep}{5pt}
    \begin{tabular}{|*{9}{c|} }
            \hline
            \multicolumn{9}{|c|}{Performance for clean data training and testing as a reference: GMM: 72.8 and NMM: 77.6} \\
            \hline \hline
            \multirow{3}{*}{Type of Noise}
            & \multicolumn{8}{c|}{SNR levels for different kinds of noises} \\
            \cline{2-9}
            & \multicolumn{2}{c|}{25dB} & \multicolumn{2}{c|}{20dB} & \multicolumn{2}{c|}{15dB} & \multicolumn{2}{c|}{10dB} \\
            \cline{2-9}
            & GMM & NMM & GMM & NMM & GMM & NMM & GMM & NMM\\
             \cline{1-9}
            \texttt{white} & 55.6 (17.2) & 67.1 (10.5) & 46.8 (26.0) & 60.0 (17.6) & 36.8 (36.0) & 49.4 (28.2) & 27.9 (44.9) & 37.7 (39.9)\\
            \hline
            \texttt{pink} & 59.9 (12.9) & 69.3 (8.3) & 51.9 (20.9) & 61.7 (15.9)  &  42.3 (30.5) & 48.6 (29) & 32.2 (40.6) & 33.7 (43.9) \\
            \hline
            \texttt{babble} & 65.7 (7.1) & 70.7 (6.9) & 59.3 (13.5) & 65.8 (11.8) & 49.3 (23.5) & 56.2 (21.4) & 37.4 (35.4) & 42.3 (35.3) \\
            \hline
            \texttt{hfchannel} & 62.3 (10.5) & 67.9 (9.7) & 54.4 (18.4) & 63.4 (14.2) & 44.1 (28.7) & 55.8 (21.8) & 33.3 (39.5) & 44.9 (32.7) \\
            \hline
            \hline
        \end{tabular}
    \label{tab:cl_tr_no_te_table_all}
\end{table*}

\subsubsection{Clean training and noisy testing - robustness test}
\label{sec:results2}
The next experiment is for checking the robustness of the NMM-HMM relative to GMM-HMM. We train using clean data and test using noisy data. We used four types of noise: white, pink, babble and high frequency channel (labelled as 'hfchannel') at different signal-to-noise-ratio (SNR) levels. The performance of a robust system is expected to deteriorate relatively slowly as the noise power increases (SNR decreases). 

Table \ref{tab:cl_tr_no_te_table_all} shows the results. The performance drop is counted with respect to the clean data performance (as seen in Table \ref{tab:NMM-HMM_diff_components} of the previous subsection~\ref{sec:results1}, and shown in parenthesis in  Table \ref{tab:cl_tr_no_te_table_all}). We find that the performance drop is gradual with decrease in SNR for white, babble and hfchannel noises. While the performance drop is gradual for NMM-HMM in case of pink noise at SNR = 25, 20 and 15 dB, there is a drastic drop for NMM-HMM when the pink noise is at 10 dB SNR. Overall the NMM-HMM shows a significantly greater noise robustness compared to GMM-HMM.   


\subsubsection{Noisy training and noisy testing - robustness test}
In this set of experiments, the models were trained using a joint dataset comprised of clean data and data corrupted with white noise at 10dB SNR. Test results are shown in Table~\ref{tab:no_tr_no_te_table_white}. We again find that NMM-HMM is more robust than GMM-HMM.

\begin{table}[!htbp]
    \centering
    \caption{Test accuracy (in \%) using noisy training. The performance drop is shown in parenthesis with respect to the clean train and clear test scenario.}
    \begin{tabular}{|c|c|c|c|}
            \hline
            \multirow{2}{*}{Model-Type} &
            \multirow{2}{*}{Clean} &
            \multicolumn{2}{c|}{\texttt{white} noise}\\
            \cline{3-4}
            &  & 15dB & 10dB \\
            \cline{1-4}
            GMM-HMM & 72.0 & 55.9 (16.1) & 53.7 (18.3) \\
            \hline
            NMM-HMM & 76.8 & 69.2 (7.6) & 65.7 (11.1) \\
            \hline
            \hline
        \end{tabular}
    \label{tab:no_tr_no_te_table_white}
\end{table}
\section{Conclusion}
\label{sec:conclusion}

In this work, we demonstrated that it is possible to use neural networks for improving maximum-likelihood based classification performance and robustness against noise. In addition the methods are able to use time-tested signal processing based features as MFCCs and machine learning techniques as expectation-maximization for training the models. In the future we can consider use of the input features such as logarithm of the power spectrum. 



\bibliographystyle{IEEEbib}
\bibliography{strings,refs}

\end{document}